\documentclass{article}

\usepackage[final]{neurips_2019}
\usepackage{graphicx}
\usepackage[utf8]{inputenc} 
\usepackage{booktabs}       
\usepackage{natbib}
\setcitestyle{square, comma, numbers,sort&compress, super}
\usepackage{etoolbox}

\makeatletter
\patchcmd{\@maketitle}{\raggedright}{\centering}{}{}
\patchcmd{\@maketitle}{\raggedright}{\centering}{}{}
\makeatother

\title{Split Learning for collaborative deep learning in healthcare}
{\centering
\author{\parbox{10cm}{
    {Maarten G.Poirot$^1$, Praneeth Vepakomma$^2$, Ken Chang$^3$, Jayashree Kalpathy-Cramer$^3$, Rajiv Gupta$^1$, Ramesh Raskar$^2$}\\
    {\normalsize
    $^1$ Department of Radiology, Massachusetts General Hospital\\
  \texttt{maartenpoirot@gmail.com} \\
   $^2$   Media Lab, Massachusetts Institute of Technology\\\texttt{ vepakom@mit.edu} \\
  $^3$ Athinoula A. Martinos Center for Biomedical Imaging, Department of Radiology, Massachusetts General Hospital \\  \texttt{kenchang@mit.edu} }}
}}

\begin{document}

\maketitle

\begin{abstract}
    Shortage of labeled data has been holding the surge of deep learning in healthcare back, as sample sizes are often small, patient information cannot be shared openly, and multi-center collaborative studies are a burden to set up. Distributed machine learning methods promise to mitigate these problems. We argue for a split learning based approach and apply this distributed learning method for the first time in the medical field to compare performance against (1) centrally hosted and (2) non collaborative configurations for a range of participants.
   \par Two medical deep learning tasks are used to compare split learning to conventional single and multi center approaches: a binary classification problem of a data set of 9000 fundus photos, and multi-label classification problem of a data set of 156,535 chest X-rays. The several distributed learning setups are compared for a range of 1-50 distributed participants. \par Performance of the split learning configuration remained constant for any number of clients compared to a single center study, showing a marked difference compared to the non collaborative configuration after 2 clients ($p<0.001$) for both sets.
    
    Our results affirm the benefits of collaborative training of deep neural networks in health care. Our work proves the significant benefit of distributed learning in healthcare, and paves the way for future real-world implementations.
    
\end{abstract}

\section{Introduction}
Deep neural networks have become the state-of-the-art for a range of tasks such as image classification, speech recognition, natural language processing\cite{Collobert:2008:UAN:1390156.1390177} and based on complex data such as electronic health records (EHR), imaging, bio-sensors, omics and text.

Learning with these networks relies on vast amounts of structured training data to achieve proper performance such that it increases generalization and robustness \cite{10.1093/bib/bbx044,Panch2018}. However, medical sample sizes tend to be small, especially in rarer diseases \cite{Dluhos2017}. Thus traditionally, clinical models have often been trained on small data sets \cite{Panch2018}.

Multi-center distributed studies can significantly increase the available amount of data and its diversity by centralization of the data sets, but it comes with several drawbacks: Setting up a multi-center organizational collaboration can be difficult as patient data can often not leave the premise due to ethical or regulatory concerns such as HIPAA \cite{Annas2003,Mercuri:2004:HHC:1005817.1005840,Nass2009,Luxton2012}. Secondly, institutions might find their data to be too valuable to share \cite{Xia2018}. Lastly, the additional storage and bandwidth required to store this data can be a burden. These factors heavily impede collaboration in health in a traditional setting.

An alternative to centrally hosting information are secure and private distributed learning solutions. 
These methods include model averaging\cite{DBLP:journals/corr/SuC15}, large scale synchronous gradient descent (LS-SGD)\cite{DBLP:journals/corr/ChenMBJ16}, federated learning\cite{DBLP:journals/corr/McMahanMRA16}, cyclical weight transfer\cite{Chang2018} and split learning\citep{GUPTA20181,Vepakomma2019Leak,SNNforhealth,singh2019detailed,sharma2019expertmatcher}. These models can be compared on several properties, which are performance with respect to a centralized setup, privacy, bandwidth usage and distribution of computational load.

From previous survey by Vepakomma et. al.\cite{DBLP:journals/corr/abs-1812-03288} several properties of these methods can be identified and weighed for our clinical implementation. Model averaging and LS-SGD only allow for synchronous training, meaning the model can only continue training after all clients have yielded their input. This would present major logistical challenges, especially when clients work with different network connection speeds or hardware configurations. Other methods like cyclical weight transfer do not preserve optimal performance compared to  computing in a centralized setting by design. Lastly, every method differs in the amount of information it reveals. For a more thorough comparison of these methods we would like to refer to the aforementioned papers. 

In this study, split learning is applied in the medical field for the first time to our knowledge. Two data sets are used: retinal fundus photos and chest X-rays. For these two data sets performance of a split learning configuration is compared to (1) centrally hosted and (2) non collaborative configuration for a range of number of distributed participants.

\section{Related work}

The concept of split learning was first introduced by \cite{GUPTA20181}. In comparisons on the CIFAR 10 and CIFAR 100 data sets; split learning has shown to outperform federated learning and LS-SGD in terms of convergence for accuracy and client side computational requirements \cite{SNNforhealth}. In addition, split learning shows improved security by reducing leakage of information as shown by Vepakomma et al \cite{Vepakomma2019Leak}. 

The paradigm of split learning revolves around splitting up a conventional neural network into several elements that can have different accessibility properties. These elements are `\textit{links}', that together form a `\textit{chain}', making up the full network. The mentioned accessibility properties of these links can either be `\textit{central}', which means they are hosted on the central server location and accessible as black box to all clients, or `\textit{local}', in which case they can only be accessed by the proprietary client. \par
\begin{figure}[h]
  \centering
  \includegraphics[width=0.7\textwidth]{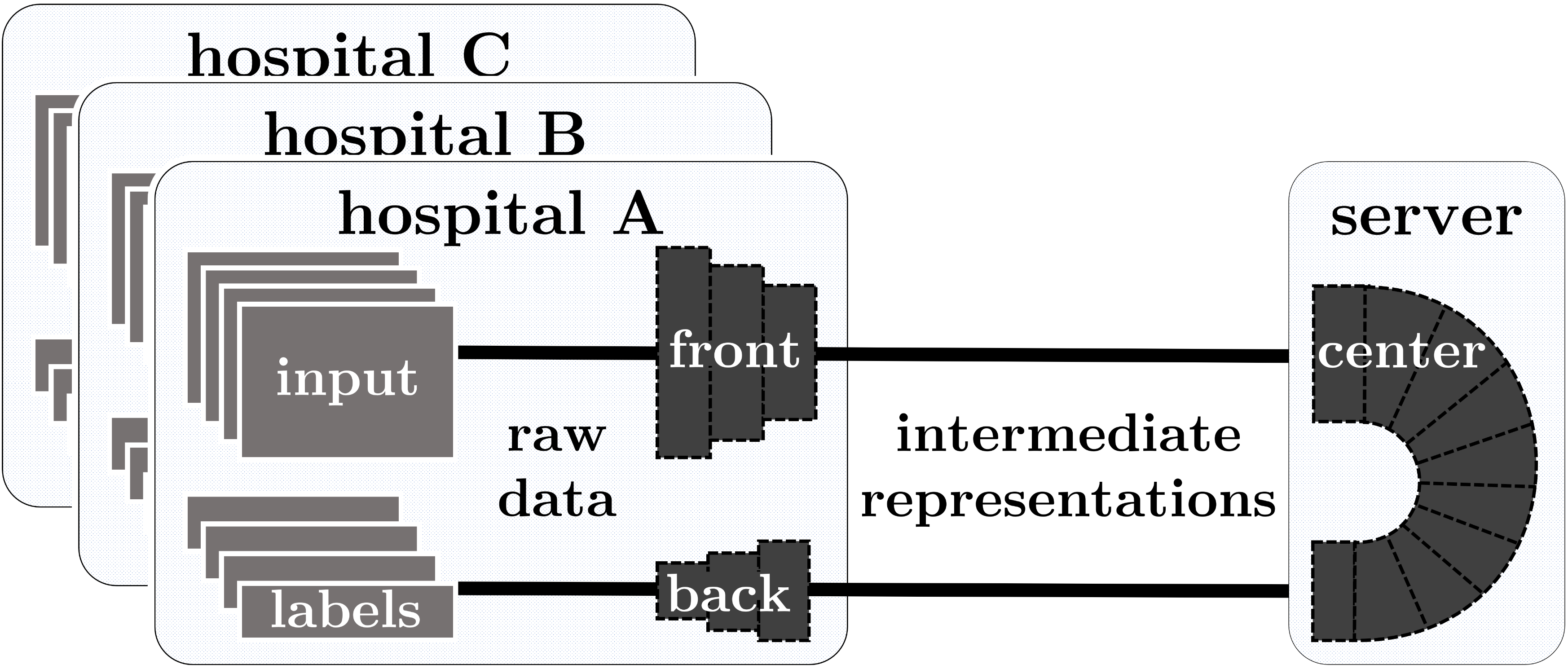}
  \caption{Graphical representation of the U-shaped configuration. Three clients named hospital A, B and C hold their own data and train a collaborative model without sharing raw data.}
  \label{fig:diagram}
\end{figure}

 \textbf{U-shaped split learning:} Although the configuration could potentially take many forms, a particular configuration called the U-shaped configuration \cite{GUPTA20181,SNNforhealth} is implemented in this work for its simplicity and suitability for healthcare. This configuration requires no raw data sharing as well as no label sharing. The chain in this configuration consists of three links. As considered from a forward propagation point of view, the first is called `\textit{front}', and is local. It receives raw input data during forward propagation, and returns an obfuscated intermediate representation. The second link is called `\textit{center}' and is centrally hosted. It takes the intermediate representation from the front, and performs most of the computation to return another intermediate representation to the final link called `\textit{back}'. The back is again local and performs the final decoding computation on its input. This local stage is where gradients are computed from the decoded output and labels. This configuration is visualized in figure \ref{fig:diagram}.

When training the model one or more mini-batches can  iteratively be forwarded through the chain thereby training both the local, as well as the central links. When training is switched from one client to another, the state of the local links from one clients is downloaded and updated at the next. The system is not dependant on results from all clients to push an update, which resolves the logistical challenges in synchronous training methods mentioned earlier.

Typically but not necessarily, the largest part of trainable networks layers can be found in the central link. This reduces bandwidth used in sharing local states, as well as client side computational cost. This property allows for computation of more complex networks for clients with less computational power, compared to federated learning.

\section{Methods}
\subsection{Data collection}
We utilized the diabetic retinopathy (DR) dataset as previously described in work by Chang et al \cite{Chang2018}. This data set originates from the Kaggle Diabetic Retinopathy dataset \cite{KaggleInc.2015} of retinal fundus photos. A subset of 9000 images was used for training and validation to prevent saturation of learning for models when trained non-collaboratively. The original multi-class classification problem was simplified to binary classification of `normal' and `abnormal'. Images were downsampled to 256x256 RGB images. For further preprocessing details we refer to Chang et al \cite{Chang2018}.

The second data set used was the large chest X-ray dataset `\textit{CheXpert}' \cite{DBLP:journals/corr/abs-1901-07031}. The data set consists of 224,316 chest radiographs with labels of 65,240 patients. The problem is posed as a multi-label classification problem of 14 common chest radiographic observations. Cases where labels contained uncertainty were excluded according to the baseline approach as described in the paper. To further decrease the data set size to prevent saturation in non-collaborative setting, some subsets of images of different shapes that were most commonly occurring (320x390 px) were excluded. This resulted in a remaining dataset of 156,535 chest radiographs.

Both data sets were partitioned to cohorts of 75\% training, and 25\% validation data.  When the data sets were further split over multiple clients, training data was split equally. Both partitioning operations are performed randomly without patients overlapping in both cohorts. Validation data was not split so as to retain the validation process even when data is split over many clients. 

\subsection{Neural networks}
For the DR dataset an implementation largely influenced by Chang et al \cite{Chang2018} was employed. A 34-layer residual network (Resnet-34)\cite{He2016DeepRL} architecture was utilized with Glorot uniform initialization \cite{Glorot10understandingthe}. Adam \cite{kingma2014adam} optimization using standard parameters ($\beta_1 = 0.9$ and $beta_2 = 0.999$), and default learning rate ($10^{-4}$) without decay was used.  Data was augmented in real-time using random rotations (0-360 degrees) and 50\% chance of lateral or axial inversion. Loss was computed used a binary cross entropy loss function. Training was performed on a GeForce GTX TITAN X graphics processing unit until validation accuracy reached a plateau as defined by not decreasing for more than 30 epochs. 

For the CheXpert dataset, the implementation as described by Rajpurkar et al \cite{DBLP:journals/corr/abs-1711-05225} was used. A 121-layer dense network (DenseNet121) \cite{DBLP:journals/corr/HuangLW16a} was pretrained on ImageNet\cite{5206848}. Loss was defined computed using a combined sigmoid binary cross entropy loss. Adam optimization using standard parameters ($\beta_1 = 0.9$ and $\beta_2 = 0.999$), and default learning rate ($10^{-4}$) without decay was used. Batch size used was 24. Data was augmented by 50\% chance of lateral inversion. Models were trained until validation loss reached a plateau as defined by not decreasing for more than five epochs. The model with the lowest validation loss was used picked. Training was performed on a Nvidia GeForce GTX 1080 Ti graphics processing unit.

In collaborative mode, every client sequentially trained the network for one epoch. Whenever training switched from one client to the next local client states were copied to next client. In non collaborative mode, a single client was trained on the same sample size of data as it would have had in the collaborative setting.

\subsection{Performance analysis}
Performance of the DR data set is defined as the highest classification accuracy on the validation set by averaging all clients. For the CheXpert data set, the receiver operating characteristic curves (ROC) were generated from the validation set, for the model state of the client with lowest loss across all mini-batches in the epoch achieving the lowest loss. Final result was the average area under the ROC (AUROC) as shown in Figure 2 across all five competition tasks as defined by the original study (Atelectasis, cardiomegaly, consolidation, edema and pleural effusion).

\section{Results}
\begin{figure}[!htbp]
  \centering
  \includegraphics[width=\textwidth]{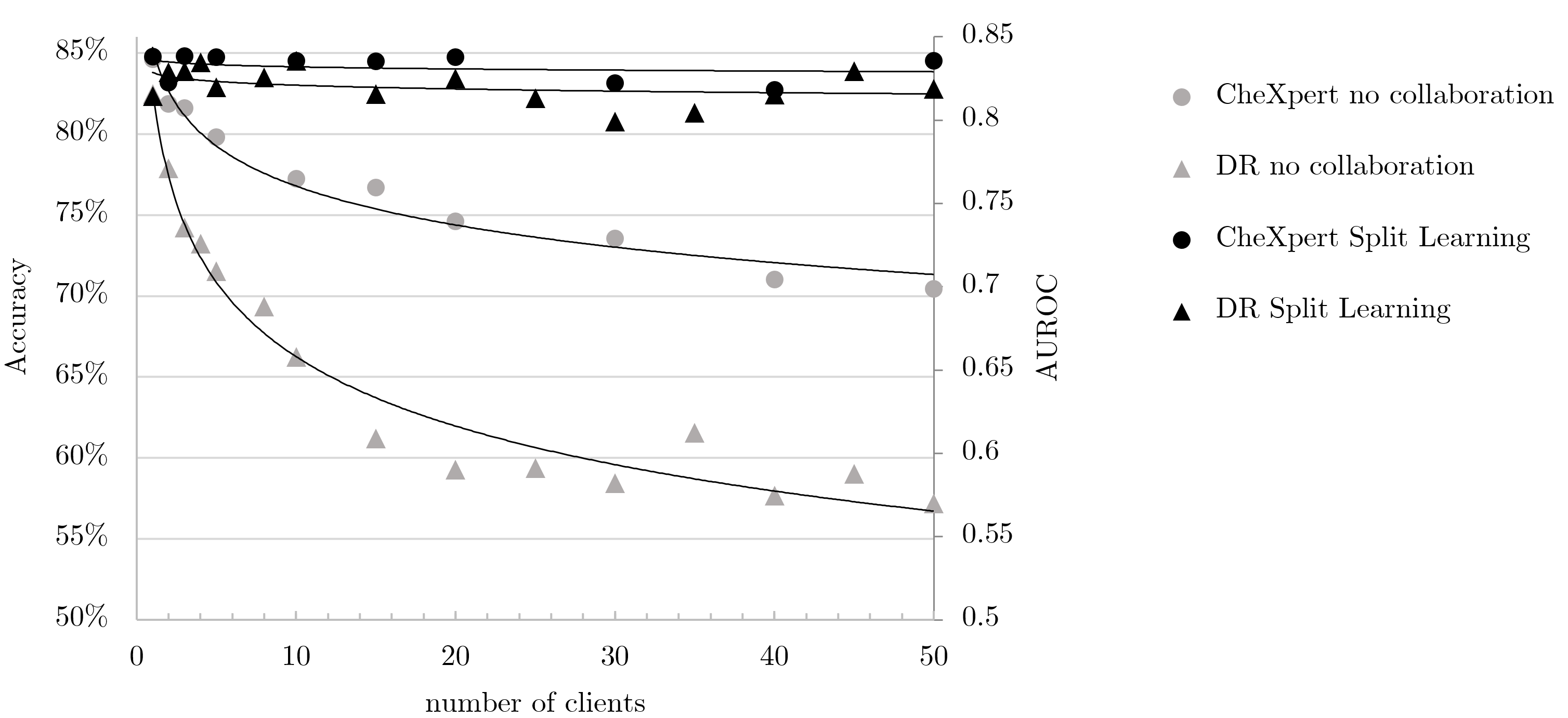}
  \caption{Performance of non collaborative (gray) and split learning (black) configurations. Number of clients refers to the number of clients the data was divided over. As the total amount of data remained constant, this directly relates to each client's sample size.}
  \label{fig:graph}
\end{figure}

The performance of split learning based configurations is compared to a non collaborative configurations for the DR data set using accuracy, and CheXpert using the AUROC, in figure \ref{fig:graph}. As shown in the figure, the split learning based approaches on both the CheXpert and diabetic retinopathy datasets performed exceedingly better than performance in non-collaborative settings \cite{DBLP:journals/corr/abs-1901-07031}. Experimental results, including bootstrapping results for the diabetic retinopathy set are given in table \ref{sample-table}. On the Chexpert split learning dataset
mean performance was significantly  ($\alpha = 0.005$) lower in non collaborative compared to collaborative setting especially in cases with $>2$ clients and two sample two tailed T-test was also used to compare means to reach this conclusion.

\begin{table}[!b]
  \caption{Performance for diabetic retinopathy }
  \label{sample-table}
  \centering
  \begin{tabular}{lll}
    \toprule
         & Split learning     & Non collaborative  \\
    number of clients & b mean & (C.I.) mean (C.I.) \\
    \midrule
 1 & 0.888 (0.896, 0.880) & 0.869 (0.877, 0.861)  \\
 2 & 0.850 (0.857, 0.843) & 0.852 (0.865, 0.839)  \\
 3 & 0.868 (0.875, 0.861) & 0.753 (0.766, 0.742)  \\
 4 & 0.884 (0.891, 0.878) & 0.754 (0.770, 0.739)  \\
 5 & 0.869 (0.877, 0.861) & 0.755 (0.772, 0.738)  \\
 8 & 0.887 (0.894, 0.880) & 0.717 (0.733, 0.701)  \\
10 & 0.858 (0.868, 0.849) & 0.676 (0.695, 0.657)  \\
15 & 0.838 (0.848, 0.829) & 0.627 (0.649, 0.603)  \\
20 & 0.860 (0.868, 0.852) & 0.613 (0.632, 0.594)  \\
25 & 0.850 (0.858, 0.841) & 0.607 (0.627, 0.588)  \\
30 & 0.814 (0.831, 0.797) & 0.620 (0.648, 0.590)  \\
35 & 0.798 (0.819, 0.780) & 0.633 (0.656, 0.611)  \\
40 & 0.852 (0.859, 0.844) & 0.595 (0.619, 0.568)  \\
45 & 0.883 (0.891, 0.876) & 0.608 (0.634, 0.581)  \\
50 & 0.859 (0.869, 0.849) & 0.588 (0.611, 0.565)  \\

    \bottomrule
  \end{tabular}
\end{table}

\section{Discussion and future work}

    Distributed machine learning based solutions can provide great benefit to the medical field by enhancing seamless collaboration across entities.
    Split learning has shown benefits compared to alternative distributed learning methods. We have applied split learning in the medical field for the first time and it worked great compared to conventional single and multi-institution setups.
    Our results show that teaming up in general can give a great performance boost. Our results show that teaming up in a distributed learning setting in general can give a great performance boost in comparison to non-collaboration. In the future we will also compare to federated learning and LS-SGD within the medical setup. These comparisons have already been made recently in the non-medical settings in \citep{GUPTA20181, Vepakomma2019Leak}. We could investigate alternative weight transfer protocols to aim to improve efficiency. We also plan to investigate privacy enhancements and alternative configurations in \cite{SNNforhealth} for healthcare settings via controlled real-world healthcare deployments.


\bibliographystyle{plainnat}

\bibliography{neurips_2019}

\begin{thebibliography}{27}
\providecommand{\natexlab}[1]{#1}
\providecommand{\url}[1]{\texttt{#1}}
\expandafter\ifx\csname urlstyle\endcsname\relax
  \providecommand{\doi}[1]{doi: #1}\else
  \providecommand{\doi}{doi: \begingroup \urlstyle{rm}\Url}\fi

\bibitem[Annas(2003)]{Annas2003}
George~J. Annas.
\newblock {HIPAA regulations - A new era of medical-record privacy?}
\newblock \emph{New England Journal of Medicine}, 348\penalty0 (15):\penalty0
  1486--1490, 2003.
\newblock ISSN 00284793.
\newblock \doi{10.1056/NEJMlim035027}.

\bibitem[Chang et~al.(2018)Chang, Balachandar, Lam, Yi, Brown, Beers, Rosen,
  Rubin, and Kalpathy-Cramer]{Chang2018}
Ken Chang, Niranjan Balachandar, Carson Lam, Darvin Yi, James Brown, Andrew
  Beers, Bruce Rosen, Daniel~L Rubin, and Jayashree Kalpathy-Cramer.
\newblock {Distributed deep learning networks among institutions for medical
  imaging.}
\newblock \emph{Journal of the American Medical Informatics Association :
  JAMIA}, 25\penalty0 (8):\penalty0 945--954, aug 2018.
\newblock ISSN 1527-974X (Electronic).
\newblock \doi{10.1093/jamia/ocy017}.

\bibitem[Chen et~al.(2016)Chen, Monga, Bengio, and
  J{\'{o}}zefowicz]{DBLP:journals/corr/ChenMBJ16}
Jianmin Chen, Rajat Monga, Samy Bengio, and Rafal J{\'{o}}zefowicz.
\newblock {Revisiting Distributed Synchronous {\{}SGD{\}}}.
\newblock \emph{CoRR}, abs/1604.00981, 2016.
\newblock URL \url{http://arxiv.org/abs/1604.00981}.

\bibitem[Collobert and Weston(2008)]{Collobert:2008:UAN:1390156.1390177}
Ronan Collobert and Jason Weston.
\newblock {A Unified Architecture for Natural Language Processing: Deep Neural
  Networks with Multitask Learning}.
\newblock In \emph{Proceedings of the 25th International Conference on Machine
  Learning}, ICML '08, pages 160--167, New York, NY, USA, 2008. ACM.
\newblock ISBN 978-1-60558-205-4.
\newblock \doi{10.1145/1390156.1390177}.
\newblock URL \url{http://doi.acm.org/10.1145/1390156.1390177}.

\bibitem[Deng et~al.(2009)Deng, Dong, Socher, Li, {Kai Li}, and {Li
  Fei-Fei}]{5206848}
J~Deng, W~Dong, R~Socher, L~Li, {Kai Li}, and {Li Fei-Fei}.
\newblock {ImageNet: A large-scale hierarchical image database}.
\newblock In \emph{2009 IEEE Conference on Computer Vision and Pattern
  Recognition}, pages 248--255, jun 2009.
\newblock \doi{10.1109/CVPR.2009.5206848}.

\bibitem[Dluhos et~al.(2017)Dluhos, Schwarz, Cahn, van Haren, Kahn, Spaniel,
  Horacek, Kasparek, and Schnack]{Dluhos2017}
Petr Dluhos, Daniel Schwarz, Wiepke Cahn, Neeltje van Haren, Rene Kahn, Filip
  Spaniel, Jiri Horacek, Tomas Kasparek, and Hugo Schnack.
\newblock {Multi-center machine learning in imaging psychiatry: A meta-model
  approach.}
\newblock \emph{NeuroImage}, 155:\penalty0 10--24, jul 2017.
\newblock ISSN 1095-9572 (Electronic).
\newblock \doi{10.1016/j.neuroimage.2017.03.027}.

\bibitem[Glorot and Bengio(2010)]{Glorot10understandingthe}
Xavier Glorot and Yoshua Bengio.
\newblock Understanding the difficulty of training deep feedforward neural
  networks.
\newblock In \emph{In Proceedings of the International Conference on Artificial
  Intelligence and Statistics (AISTATS’10). Society for Artificial
  Intelligence and Statistics}, 2010.

\bibitem[Gupta and Raskar(2018)]{GUPTA20181}
Otkrist Gupta and Ramesh Raskar.
\newblock {Distributed learning of deep neural network over multiple agents}.
\newblock \emph{Journal of Network and Computer Applications}, 116:\penalty0
  1--8, 2018.
\newblock ISSN 1084-8045.
\newblock \doi{https://doi.org/10.1016/j.jnca.2018.05.003}.
\newblock URL
  \url{http://www.sciencedirect.com/science/article/pii/S1084804518301590}.

\bibitem[He et~al.(2016)He, Zhang, Ren, and Sun]{He2016DeepRL}
Kaiming He, Xiangyu Zhang, Shaoqing Ren, and Jian Sun.
\newblock {Deep Residual Learning for Image Recognition}.
\newblock \emph{2016 IEEE Conference on Computer Vision and Pattern Recognition
  (CVPR)}, pages 770--778, 2016.

\bibitem[Huang et~al.(2016)Huang, Liu, and
  Weinberger]{DBLP:journals/corr/HuangLW16a}
Gao Huang, Zhuang Liu, and Kilian~Q Weinberger.
\newblock {Densely Connected Convolutional Networks}.
\newblock \emph{CoRR}, abs/1608.06993, 2016.
\newblock URL \url{http://arxiv.org/abs/1608.06993}.

\bibitem[Irvin et~al.(2019)Irvin, Rajpurkar, Ko, Yu, Ciurea{-}Ilcus, Chute,
  Marklund, Haghgoo, Ball, Shpanskaya, Seekins, Mong, Halabi, Sandberg, Jones,
  Larson, Langlotz, Patel, Lungren, and Ng]{DBLP:journals/corr/abs-1901-07031}
Jeremy Irvin, Pranav Rajpurkar, Michael Ko, Yifan Yu, Silviana Ciurea{-}Ilcus,
  Chris Chute, Henrik Marklund, Behzad Haghgoo, Robyn~L. Ball, Katie~S.
  Shpanskaya, Jayne Seekins, David~A. Mong, Safwan~S. Halabi, Jesse~K.
  Sandberg, Ricky Jones, David~B. Larson, Curtis~P. Langlotz, Bhavik~N. Patel,
  Matthew~P. Lungren, and Andrew~Y. Ng.
\newblock Chexpert: {A} large chest radiograph dataset with uncertainty labels
  and expert comparison.
\newblock \emph{CoRR}, abs/1901.07031, 2019.
\newblock URL \url{http://arxiv.org/abs/1901.07031}.

\bibitem[{Kaggle Inc.}(2015)]{KaggleInc.2015}
{Kaggle Inc.}
\newblock {Diabetic Retinopathy Detection | Kaggle}, 2015.
\newblock URL \url{https://www.kaggle.com/c/diabetic-retinopathy-detection}.

\bibitem[Kingma and Ba(2014)]{kingma2014adam}
Diederik~P Kingma and Jimmy Ba.
\newblock {Adam: A method for stochastic optimization}.
\newblock \emph{arXiv preprint arXiv:1412.6980}, 2014.

\bibitem[Luxton et~al.(2012)Luxton, Kayl, and Mishkind]{Luxton2012}
David~D Luxton, Robert~A Kayl, and Matthew~C Mishkind.
\newblock {mHealth data security: the need for HIPAA-compliant
  standardization.}
\newblock \emph{Telemedicine journal and e-health : the official journal of the
  American Telemedicine Association}, 18\penalty0 (4):\penalty0 284--288, may
  2012.
\newblock ISSN 1556-3669 (Electronic).
\newblock \doi{10.1089/tmj.2011.0180}.

\bibitem[McMahan et~al.(2016)McMahan, Moore, Ramage, and
  y~Arcas]{DBLP:journals/corr/McMahanMRA16}
H.~Brendan McMahan, Eider Moore, Daniel Ramage, and Blaise~Ag{\"{u}}era
  y~Arcas.
\newblock Federated learning of deep networks using model averaging.
\newblock \emph{CoRR}, abs/1602.05629, 2016.
\newblock URL \url{http://arxiv.org/abs/1602.05629}.

\bibitem[Mercuri(2004)]{Mercuri:2004:HHC:1005817.1005840}
Rebecca~T Mercuri.
\newblock {The HIPAA-potamus in Health Care Data Security}.
\newblock \emph{Commun. ACM}, 47\penalty0 (7):\penalty0 25--28, jul 2004.
\newblock ISSN 0001-0782.
\newblock \doi{10.1145/1005817.1005840}.
\newblock URL \url{http://doi.acm.org/10.1145/1005817.1005840}.

\bibitem[Miotto et~al.(2017)Miotto, Wang, Wang, Jiang, and
  Dudley]{10.1093/bib/bbx044}
Riccardo Miotto, Fei Wang, Shuang Wang, Xiaoqian Jiang, and Joel~T Dudley.
\newblock {Deep learning for healthcare: review, opportunities and challenges}.
\newblock \emph{Briefings in Bioinformatics}, 19\penalty0 (6):\penalty0
  1236--1246, 2017.
\newblock ISSN 1477-4054.
\newblock \doi{10.1093/bib/bbx044}.
\newblock URL \url{https://doi.org/10.1093/bib/bbx044}.

\bibitem[Nass et~al.(2009)Nass, Levit, and Gostin]{Nass2009}
Sharyl~J. Nass, Laura~A. Levit, and Lawrence~O. Gostin.
\newblock \emph{{Beyond the HIPAA Privacy Rule: Enhancing Privacy, Improving
  Health Through Research}}.
\newblock National Academies Press, 2009.
\newblock ISBN 9780309141376.
\newblock \doi{10.17226/12458}.

\bibitem[Panch et~al.(2018)Panch, Szolovits, and Atun]{Panch2018}
Trishan Panch, Peter Szolovits, and Rifat Atun.
\newblock {Artificial intelligence, machine learning and health systems.}
\newblock \emph{Journal of global health}, 8\penalty0 (2):\penalty0 20303, dec
  2018.
\newblock ISSN 2047-2986 (Electronic).
\newblock \doi{10.7189/jogh.08.020303}.

\bibitem[Rajpurkar et~al.(2017)Rajpurkar, Irvin, Zhu, Yang, Mehta, Duan, Ding,
  Bagul, Langlotz, Shpanskaya, Lungren, and
  Ng]{DBLP:journals/corr/abs-1711-05225}
Pranav Rajpurkar, Jeremy Irvin, Kaylie Zhu, Brandon Yang, Hershel Mehta, Tony
  Duan, Daisy~Yi Ding, Aarti Bagul, Curtis Langlotz, Katie~S. Shpanskaya,
  Matthew~P. Lungren, and Andrew~Y. Ng.
\newblock Chexnet: Radiologist-level pneumonia detection on chest x-rays with
  deep learning.
\newblock \emph{CoRR}, abs/1711.05225, 2017.
\newblock URL \url{http://arxiv.org/abs/1711.05225}.

\bibitem[Sharma et~al.(2019)Sharma, Vepakomma, Swedish, Chang, Kalpathy-Cramer,
  and Raskar]{sharma2019expertmatcher}
Vivek Sharma, Praneeth Vepakomma, Tristan Swedish, Ken Chang, Jayashree
  Kalpathy-Cramer, and Ramesh Raskar.
\newblock Expertmatcher: Automating ml model selection for clients using hidden
  representations.
\newblock \emph{arXiv preprint arXiv:1910.03731}, 2019.

\bibitem[Singh et~al.(2019)Singh, Vepakomma, Gupta, and
  Raskar]{singh2019detailed}
Abhishek Singh, Praneeth Vepakomma, Otkrist Gupta, and Ramesh Raskar.
\newblock Detailed comparison of communication efficiency of split learning and
  federated learning.
\newblock \emph{arXiv preprint arXiv:1909.09145}, 2019.

\bibitem[Su and Chen(2015)]{DBLP:journals/corr/SuC15}
Hang Su and Haoyu Chen.
\newblock Experiments on parallel training of deep neural network using model
  averaging.
\newblock \emph{CoRR}, abs/1507.01239, 2015.
\newblock URL \url{http://arxiv.org/abs/1507.01239}.

\bibitem[Vepakomma et~al.(2017)Vepakomma, Gupta, Dubey, and
  Raskar]{Vepakomma2019Leak}
Praneeth Vepakomma, Otkrist Gupta, Abhimanyu Dubey, and Ramesh Raskar.
\newblock {Reducing leakage in distributed deep learning}.
\newblock \emph{AI for social good}, pages 1--6, 2017.

\bibitem[Vepakomma et~al.(2018{\natexlab{a}})Vepakomma, Gupta, Swedish, and
  Raskar]{SNNforhealth}
Praneeth Vepakomma, Otkrist Gupta, Tristan Swedish, and Ramesh Raskar.
\newblock Split learning for health: Distributed deep learning without sharing
  raw patient data.
\newblock \emph{CoRR}, abs/1812.00564, 2018{\natexlab{a}}.
\newblock URL \url{http://arxiv.org/abs/1812.00564}.

\bibitem[Vepakomma et~al.(2018{\natexlab{b}})Vepakomma, Swedish, Raskar, Gupta,
  and Dubey]{DBLP:journals/corr/abs-1812-03288}
Praneeth Vepakomma, Tristan Swedish, Ramesh Raskar, Otkrist Gupta, and
  Abhimanyu Dubey.
\newblock No peek: {A} survey of private distributed deep learning.
\newblock \emph{CoRR}, abs/1812.03288, 2018{\natexlab{b}}.
\newblock URL \url{http://arxiv.org/abs/1812.03288}.

\bibitem[Xia et~al.(2018)Xia, Wan, Yin, Gaupp, Liu, Clayton, Kantarcioglu,
  Vorobeychik, and Malin]{Xia2018}
Weiyi Xia, Zhiyu Wan, Zhijun Yin, James Gaupp, Yongtai Liu, Ellen~Wright
  Clayton, Murat Kantarcioglu, Yevgeniy Vorobeychik, and Bradley~A Malin.
\newblock {It's all in the timing: calibrating temporal penalties for
  biomedical data sharing.}
\newblock \emph{Journal of the American Medical Informatics Association :
  JAMIA}, 25\penalty0 (1):\penalty0 25--31, jan 2018.
\newblock ISSN 1527-974X (Electronic).
\newblock \doi{10.1093/jamia/ocx101}.

\end{thebibliography}

\end{document}